%% file: main.tex
\documentclass[10pt,twocolumn,letterpaper]{article}

\usepackage[pagenumbers]{cvpr} 

\usepackage[utf8]{inputenc} 
\usepackage[T1]{fontenc}    
\usepackage[pagebackref=true,breaklinks=true,letterpaper=true,colorlinks,bookmarks=false]{hyperref}       
\usepackage{url}            
\usepackage{booktabs}       
\usepackage{amsfonts}       
\usepackage{nicefrac}       
\usepackage{microtype}      
\usepackage{xcolor}         

\usepackage{times}
\usepackage{epsfig}
\usepackage{graphicx}
\usepackage{amsmath}
\usepackage{amssymb}

\usepackage{bm}
\usepackage{amsmath} 
\usepackage{xcolor}

\usepackage[pagebackref=true,breaklinks=true,letterpaper=true,colorlinks,bookmarks=false]{hyperref}
\usepackage[ruled,vlined]{algorithm2e}

\newcommand{\nosemic}{\renewcommand{\@endalgocfline}{\relax}}
\newcommand{\dosemic}{\renewcommand{\@endalgocfline}{\algocf@endline}}

\usepackage{graphicx}
\usepackage{amsmath}
\usepackage{amssymb}
\usepackage{booktabs}

\providecommand{\naturalnum}{\mathbb{N}}

\DeclareMathOperator*{\argmin}{{arg}\min}
\definecolor{ddgray}{gray}{0.5}
\newcommand{\algcomment}[1]{\textcolor{ddgray}{\textit{#1}}}

\usepackage[bottom]{footmisc}

\usepackage[capitalize]{cleveref}
\crefname{section}{Sec.}{Secs.}
\Crefname{section}{Section}{Sections}
\Crefname{table}{Table}{Tables}
\crefname{table}{Tab.}{Tabs.}

\begin{document}

\title{Cluster-guided Image Synthesis with Unconditional Models}

\author{Markos Georgopoulos\\
Imperial College London, UK\\
{\tt\small m.georgopoulos@imperial.ac.uk}
\and
James Oldfield\\
Queen Mary University of London, UK\\
{\tt\small j.a.oldfield@qmul.ac.uk}
\and
Grigorios G Chrysos\\
EPFL, Switzerland\\
{\tt\small grigorios.chrysos@epfl.ch}
\and
Yannis  Panagakis\\
University of Athens, Greece\\
{\tt\small yannisp@di.uoa.gr}
}
\maketitle


\begin{abstract}
Generative Adversarial Networks (GANs) are the driving force behind the state-of-the-art in image generation. Despite their ability to synthesize high-resolution photo-realistic images, generating content with on-demand conditioning of different granularity remains a challenge. This challenge is usually tackled by annotating massive datasets with the attributes of interest, a laborious task that is not always a viable option. Therefore, it is vital to introduce control into the generation process of unsupervised generative models. In this work, we focus on controllable image generation by leveraging GANs that are well-trained in an unsupervised fashion.
To this end, we discover that the representation space of intermediate layers of the generator forms a number of clusters that separate the data according to semantically meaningful attributes (e.g., hair color and pose).
By conditioning on the cluster assignments, the proposed method is able to control the semantic class of the generated image. 
Our approach enables sampling from each cluster by Implicit Maximum Likelihood Estimation (IMLE). We showcase the efficacy of our approach on faces (CelebA-HQ and FFHQ), animals (Imagenet) and objects (LSUN) using different pre-trained generative models. The results highlight the ability of our approach to condition image generation on attributes like gender, pose and hair style on faces, as well as a variety of features on different object classes.

\end{abstract}

\input{sections/intro}

\input{sections/related}
\input{sections/method}
\input{sections/experiments}

\input{sections/limitations}

\input{sections/conclusion}

{\small
\bibliographystyle{ieee_fullname}
\bibliography{main}
}

\end{document}

%% file: sections/intro.tex
\section{Introduction}
\label{sec:cluster_gan_intro}

Generative Adversarial Nets (GANs)~\cite{goodfellow2014generative} have demonstrated photo-realistic generation quality by utilizing the rich corpus of available image datasets. Despite their success, the value they can add as data generation tools is currently limited by the lack of control in the synthesized content. 
In the typical GAN setting an image is synthesized by sampling a vector from a latent distribution and performing a forward pass through a generator network. However, random sampling from the latent distribution provides no control over semantic attributes in the image space.
Such control over the generated characteristics is vital for tasks like autonomous driving~\cite{uvrivcavr2019yes} or (inverse) reinforcement learning~\cite{ho2016generative}.

\begin{figure*}[t!]
\centering{
  \includegraphics[width=\linewidth]{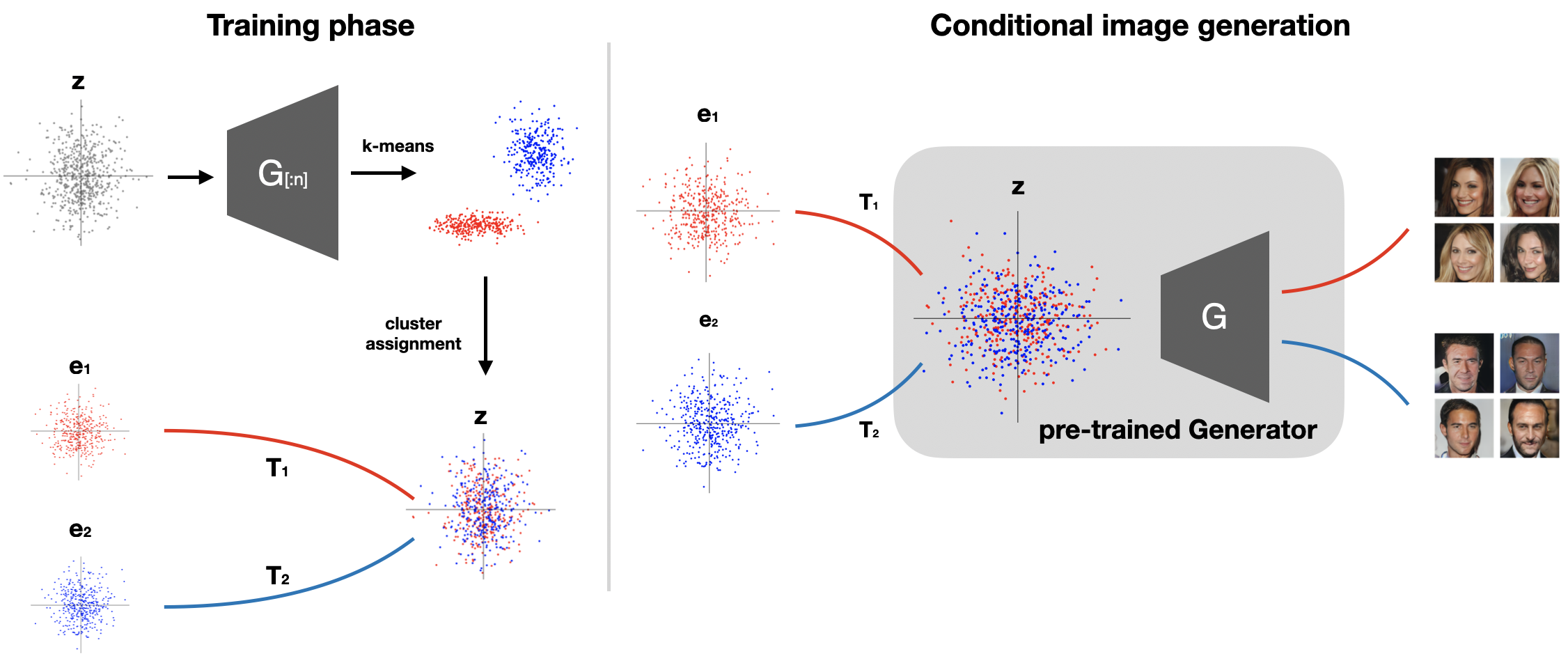}
  \caption{Conditional generation using an unsupervised generator. The training phase (depicted on the left) includes the following steps: (a) latent codes are sampled from the latent space of the generator, and then (b) passed through the first $n$ layers of the generator. The resulting representations are then clustered using k-means. Thus, we can assign each sampled latent code to a cluster (in the representation space). (c) Sequentially, we can learn a mapping from an auxiliary distribution $e_c$ to the subspace of each cluster in the latent space of the generator. In the testing phase (depicted on the right), we can sample from the auxiliary distributions and use the corresponding mappings ($T_1$ or $T_2$) to synthesize images that have specific semantic attributes, e.g., male or female. }
  
  \label{fig:motivational}
  }
\end{figure*}

A common solution to the problem of controllable generation is to introduce supervision in the form of class labels~\cite{miyato2018cgans, brock2019large, chrysos2021conditional}. This process requires the annotation of the training set, which can be a resource-intensive task, in addition to being impractical for a continually-growing number of attributes of interest. 
Additionally, even with a rich and diverse annotated dataset, training a conditional generative model that can balance control and photo-realism is a non-trivial task that requires tailored engineering tricks (e.g., truncation trick \cite{brock2019large}).

In this work, we introduce a method that can be implemented on top of any pretrained GAN to introduce control without the need for labels and supervision. The method relies on the clusters that are formed in the intermediate representation space of a generator. We posit that the representational capacity of the network allows for semantic attributes, like hair color and pose, to be disentangled in this representation space. Hence, each of the formed clusters corresponds to a different semantic attribute. This assumption enables us to control image generation by conditioning on the cluster assignment. Latent sampling from these cluster is achieved via Implicit Maximum Likelihood Estimation (IMLE). The proposed framework is summarized in Figure \ref{fig:motivational}.
We benchmark the method against GANs that learn clustering in the latent space as well as methods for interpretable directions in pretrained GANs. The results highlight the efficacy of our method in consistently generating images of desired attributes.

%% file: sections/related.tex
\section{Related work}
\label{sec:cluster_gan_related}
The related work is divided into four parts: (a) firstly, we discuss generative adversarial networks and approaches that introduce structure in the latent space. Then, (b) we refer to different methods on generative models that enable (partial) disentanglement of certain attributes. We also focus on (c) methods that learn interpretable direction in the latent space of GANs, and lastly (d) we discuss discriminative and generative approaches that utilize clustering in their latent space.

\paragraph{Generative Adversarial Nets (GANs)}~\cite{goodfellow2014generative} are able to synthesize diverse and photo-realistic images~\cite{karras2018style, brock2019large, karras2020training}. 
Introducing structure in the latent space of the generator is an active area of research. Different distributions have been proposed to enforce this structure. Specifically, a Cauchy distribution~\cite{lesniak2019distribution}, a mixture model~\cite{gurumurthy2017deligan}, a parametric distribution based on tensor decompositions~\cite{kuznetsov2019prior, georgopoulos2020multilinear}, or non-parametric distributions~\cite{singh2019non} have been used in this context. The goal is to primarily improve either the training of GANs~\cite{gurumurthy2017deligan} or the synthesized image quality~\cite{singh2019non, lesniak2019distribution}. Our method relies on a trained generator instead, hence it could utilize any of the aforementioned modifications on the latent space.

\paragraph{Disentanglement of the latent space:}  
The topic of disentangling the factors of variation in the latent space has sparked the interest of the community. InfoGAN~\cite{chen2016infogan} is the first effort to augment GAN to achieve unsupervised disentanglement. InfoGAN uses auxiliary codes $\bm{\psi}$ and maximizes the mutual information of the codes with the synthesized image. The idea has since been extended in \cite{lin2020infogan, liu2019oogan}. However, in \cite{lin2020infogan} the authors explain why the success of disentanglement relies heavily on design choices and inductive biases in the network, making ideas such as InfoGAN sensitive to the choice of the architecture. 
The works of \cite{li2018unsupervised, lee2020high, kaneko2018generative} also rely on modifying the GAN architecture with codes $\bm{\psi}$. In \cite{li2018unsupervised}, they rely on pairwise differences between elements $\psi_i$; in \cite{lee2020high} a beta-VAE~\cite{higgins2017beta} provides the codes $\bm{\psi}$, while in \cite{kaneko2018generative} the codes are provided by a tree-structure. 

Due to the challenging nature of unsupervised disentanglement often some (weak) labeling is used. In \cite{singh2019finegan}, the authors use bounding boxes as a weak supervision signal to disentangle the background from the foreground information in synthesized images. Supervised disentanglement has also been used in various tasks~\cite{tran2017disentangled, zhu2018visual}. However, in our work we do not utilize any type of labels for training.

\paragraph{Interpretable directions in GANs:} Beyond the aforementioned methods, the discovery of interpretable latent directions in a pretrained generator is an active area of study. A dataset of trajectories in the latent space is created in \cite{plumerault2020controlling}. Such trajectories correspond to known transformations in the data space; the method searches for simple transformations encoded. In \cite{voynov2020unsupervised}, they use two latent codes (one is shifted version of the other); the authors synthesize the two images and then learn a dense layer to predict the shift in the codes. Harkonen et al. \cite{harkonen2020ganspace} and Shen et al. \cite{shen2020closed} find the principal directions of variation using Principal Component Analysis (PCA), either in the latent space or the weights of the first layer. Similarly, Tensor Component Analysis is utilized in \cite{oldfield2021tensor} to better separate style and geometry. The drawbacks of such methods is that they provide interpretable directions relative to the input image; that is given a single latent code, they can find some directions that transform (e.g., rotate) the generated image. On the contrary, our method learns to directly sample an image with a desired attribute from noise without editing.

\paragraph{Clustering in the latent space:}
A long line of research takes advantage of the clusters that are formed in the feature space of convolutional representations. The majority of such works focus on unsupervised/self-supervised techniques for discriminative tasks, e.g., \cite{yan2020clusterfit, yang2016joint, zhuang2019local, caron2018deep, caron2020unsupervised}, while a number of works apply similar techniques to generative modelling \cite{mukherjee2019clustergan, liu2020diverse, georgopoulos2020multilinear}.
More closely related to this work, clustering has been utilized in the latent space of GANs \cite{mukherjee2019clustergan, liu2020diverse} to generate diverse images. Mukherjee et al. \cite{mukherjee2019clustergan} utilize an auxiliary encoder to predict the cluster assignments.
On the other hand, Liu et al. \cite{liu2020diverse} use the features from the discriminator to cluster the images. The proposed approach is orthogonal to these methods since it works on a pretrained generator and is not trained end-to-end.

%% file: sections/method.tex
\section{Method}

\begin{figure*}
\centering{
  \includegraphics[width=0.9\textwidth]{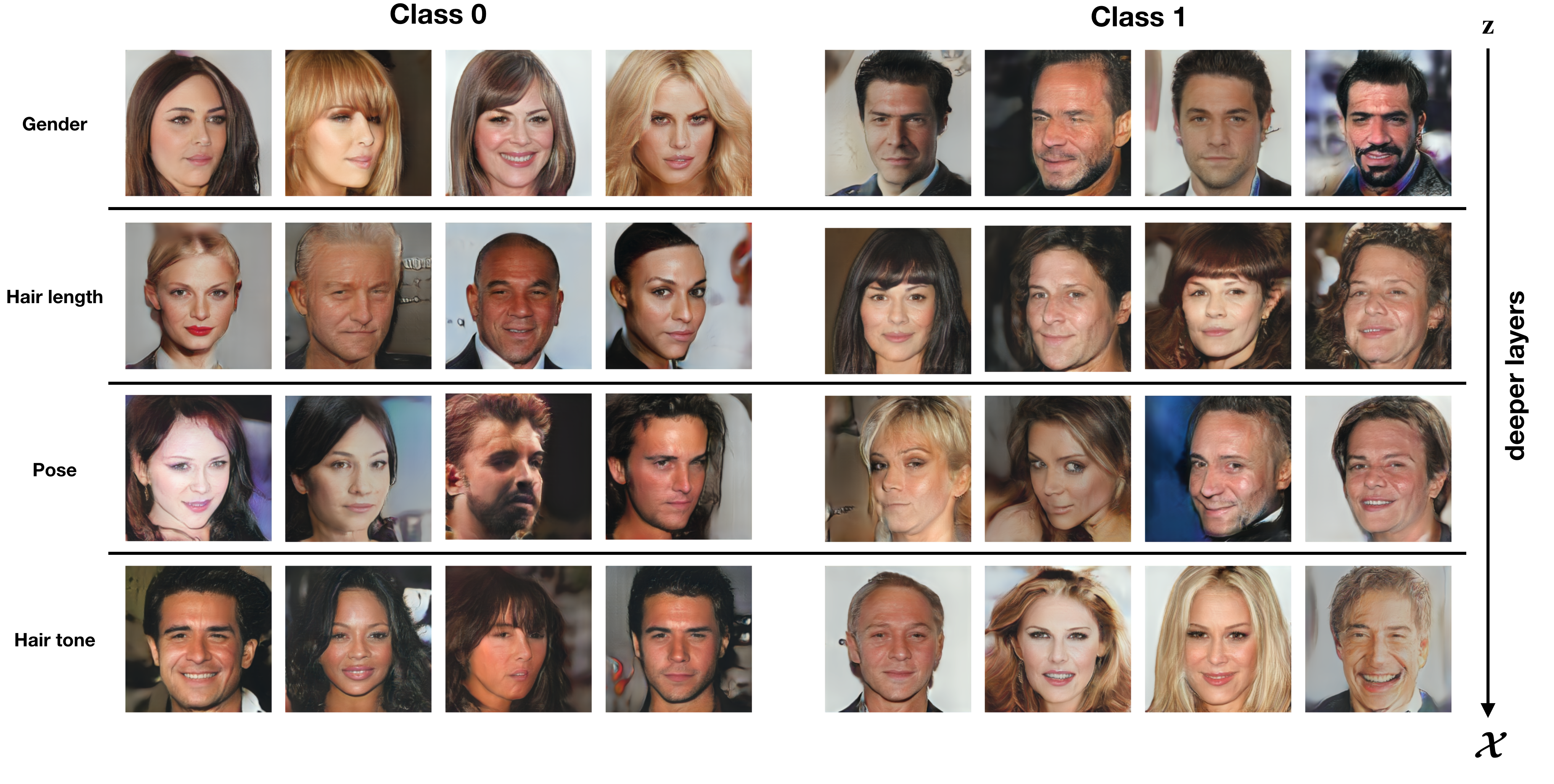}
  \caption{The mappings that are learned for clusters in different layers correspond to different semantic attributes. At each level of the PGAN generator we learn the two depicted clusters. Naturally, some of the semantic attributes are entangled, e.g., hair tone and background.}
  \label{fig:pgan_layers}
  }
\end{figure*}

In this section, we motivate our approach and present the proposed method that enables conditional generation using GANs that are not trained with attribute supervision. First, we provide a short introduction to image synthesis using GANs. We continue by motivating our assumption regarding the clustering that occurs in the representation space. Lastly, we present our method for performing conditional image generation using IMLE.

\subsection{Image synthesis with GANs}

Generative Adversarial Networks (GANs)~\cite{goodfellow2014generative} utilize a set of images $\mathcal{S} = [\bm{x}_1, \bm{x}_2, \ldots, \bm{x}_N]$ to synthesize new images 
that resemble the data in $\mathcal{S}$. 
Image synthesis is enabled by sampling a latent code  $\bm{z}\sim\mathcal{N}(\bm{0},\bm{I})$, where $\bm{I}$ is the identity matrix, and passing the code though a generator network $G$. In GANs, in addition to the generator, an auxiliary ``discriminator'' network is used for the optimization. Specifically, GANs are trained using a minimax formulation between the generator $G$ and the discriminator $D$.
The discriminator is trained to distinguish the synthetic images $\bm{\tilde{x}} = G(\bm{z})$ from the real images in $\mathcal{S}$. Concretely, we denote with $p_{\mathcal{S}}$ the data distribution
and with $p_{\bm{z}}$ the distribution for sampling the latent codes (e.g., a normal distribution). Then the learning objective can be formulated as:

\begin{equation}
\begin{split}
    \min_{\bm{w}_G} \max_{\bm{w}_D} \mathcal{L}_{GAN}(G, D) = \mathbb{E}_{\bm{x} \sim p_{S}}[\log D(\bm{x}; \bm{w}_D)] + \\ 
    \mathbb{E}_{\bm{z} \sim p_{z}}[\log (1-D(G(\bm{z}; \bm{w}_G); \bm{w}_D))]
\end{split}
\end{equation}
where the $\bm{w}_G$ and the $\bm{w}_D$ are the learnable parameters of the generator and the discriminator networks respectively. After training, image synthesis is performed by sampling from $p_{\bm{z}}$ and passing the code through $G$, which progressively increases the higher frequency content at each layer.

\subsection{Clustering in the representation space}
The core operation behind state-of-the-art GANs is convolution. This is due to the inductive bias of the operator that allows for great generalization power of these networks in the image domain. This inductive bias is so effective that even randomized convolutional neural networks (CNNs) can produce useful image representations (classification accuracy of $12\%$ on Imagenet in \cite{noroozi2016unsupervised}, while random chance is at $0.1\%$). 
A number of works \cite{caron2018deep, caron2020unsupervised} focus on leveraging the clustering that occurs in the representation space of CNNs for downstream tasks. Contrary to this line of research, this work focuses on the clustering that separates the representation space of intermediate layers of generative CNNs (i.e., GANs).

In particular, we posit that images are clustered in the representation space of the generator according to semantic attributes, e.g., geometric features. In the same vein, for a generator $G$ with a hierarchical architecture (e.g., Progressive GAN~\cite{karras2017progressive}) different layers should capture different attributes. To this end, we perform clustering on the representation $G_{[:n]}(\bm{z})$ of the $n^{\text{th}}$ layer and separate the space into $C \in \naturalnum$ clusters. In this work, we use k-means, although any clustering technique would perform in a similar manner. By manually assigning a semantic attribute to each cluster, we can perform synthesis of a specific attribute by conditioning image generation on a specific cluster.

\subsection{Implicit Maximum Likelihood Estimation}

Given the latent vector $\bm{z}$, the transformation $G_{[:n]}(\bm{z})$ to the clusters in the representation space is deterministic, thus we have a direct assignment of each latent vector $\bm{z}$ into a cluster $c$ with $c \in \{1, 2, \ldots, C\}$. We denote the latent vector corresponding to cluster $c$ as  $\bm{z}^c$. The codes $\bm{z}^c$ do not form clear clusters in the latent space nor do they follow a known probability distribution. Hence, sampling from $\bm{z}^c$ to perform conditional generation is non-trivial. This step is crucial in order to effectively sample from the observed clusters in the representation space, and proceed with the forward pass to an image of defined attributes.
To this end, we utilize an auxiliary (normal) distribution $\bm{e}^c$ and obtain a mapping $\bm{e}^c \mapsto \bm{z}^c$ using Implicit Maximum Likelihood Estimation (IMLE).

IMLE is a non-adversarial method that learns a mapping $T$ between two distributions. 
Li et al.~\cite{li2018implicit} show that the method is equivalent to maximizing the likelihood under some assumptions.
We utilize IMLE to learn a mapping from the auxiliary distribution (which is known, e.g., a Gaussian distribution) to the subspace 
spanned by all the latent vectors $\bm{z}^c$ corresponding to a specific cluster $c$. 
By learning a mapping $T_c$ for every cluster $c$, we are able to sample from the auxiliary Gaussian distribution, and obtain a synthesized image with the semantic attribute of cluster $c$.   

Next, we elaborate on the training procedure for the mappings $T_c$, which includes the following steps:

\begin{enumerate}
    \item Firstly, we sample $\Gamma \in \naturalnum$ vectors $\bm{e}^{c}$ from the auxiliary distribution for cluster $c$ and apply the transformation $T_c$ to obtain the latent codes $\bm{\tilde{z}}^c$, i.e., $\bm{\tilde{z}}^c _{\gamma}= T_c(\bm{e}^{c}_{\gamma})$ for $\gamma=1, 2, \ldots, \Gamma$.
    \item For each latent vector $\bm{z}^c_{i}$, we aim to minimize the Euclidean distance of $\bm{z}^c_{i}$ and $\bm{\tilde{z}}^c_{\gamma}$, i.e., we perform a nearest neighbor search on the vectors $\bm{e}^{c}_{\gamma}$. That is expressed as:
    \begin{equation}
        \bm{e}^c_i = \argmin_{\bm{e}^{c}_{\gamma}, \gamma=1, 2, \ldots, \Gamma} \|\bm{z}^c_i - T_c(\bm{e}^{c}_{\gamma})\|^2_2.
    \end{equation}
    \item The last step consists in optimizing the mappings $T_c$. The approximate matches of the last step are used to optimize the transformations. Concretely:
    \begin{equation}
        \tilde{T_c} = \argmin_{T_c}\sum_i \|\bm{z^c_i} - T_c(\bm{e^c_i})\|^2_2.
\end{equation}
\end{enumerate}

The steps are repeated until convergence of all mappings $T_c$.

After training the mapping functions for each cluster, conditional sampling for each semantic attribute can be performed by utilizing the corresponding mapping, i.e., $G(\bm{z}, c) = G(T_c(\bm{e}^{c}))$. The training and testing phases of the proposed framework are summarized in Figure \ref{fig:motivational}.

\begin{algorithm}[h]
\SetAlgoLined
\caption{Algorithm for the proposed method}
\KwResult{A set of mappings $\{ T_1, \dots, T_C\}$}
$\bf{z} \gets \text{Sample from the latent distribution of GAN}$

$\bf{y} \gets G_{[:n]}(z)$

$\text{Initialize parameters } \theta_c \text{ of } T_c, c \in \{ 1, \dots, C\}$  

 \For{c in 1\dots C}{
 
     \For{number of epochs}{
      $\bm{e}^{c} \gets \text{Sample from the normal distribution}$ 
      
      $\hspace{2.3em} \text{for cluster } c$
      
      $\bm{z^c} \gets \text{Latent codes belonging to cluster } c$
      
        \For{$\bm{z^c_i}$ in $\bm{z^c}$}{
          $\bm{e^c_i} \gets \argmin_{\bm{e}^{c}} \|\bm{z}^c_i - T_c(\bm{e}^{c})\|^2_2$ 
          }
        \For{number of batches}{
           // \algcomment{SGD}
           
           $\theta_c  = \theta_c - \lambda_{t} \triangledown_{\theta} \text{MSE}(T_c(\bm{e}^{c}_i), \bm{z}^c_i)$. 
           }
         }
    }
\end{algorithm}

Using IMLE to learn the mapping from the auxiliary distribution to the latent codes of the generator yields a number of benefits. For example, using a GAN for this task would suffer from unstable training as well as mode-collapse. On the other hand, IMLE ensures support for every point in the training set.
However, using IMLE to generate images directly with an L2 loss would result in blurry images. We mitigate this by using the pretrained generator to synthesize high resolution photo-realistic images from the mapped latent codes.

%% file: sections/experiments.tex
\section{Experimental evaluation}
\label{sec:cluster_gan_experiments}

\begin{figure}[t]
\centering{
  \includegraphics[width=0.9\linewidth]{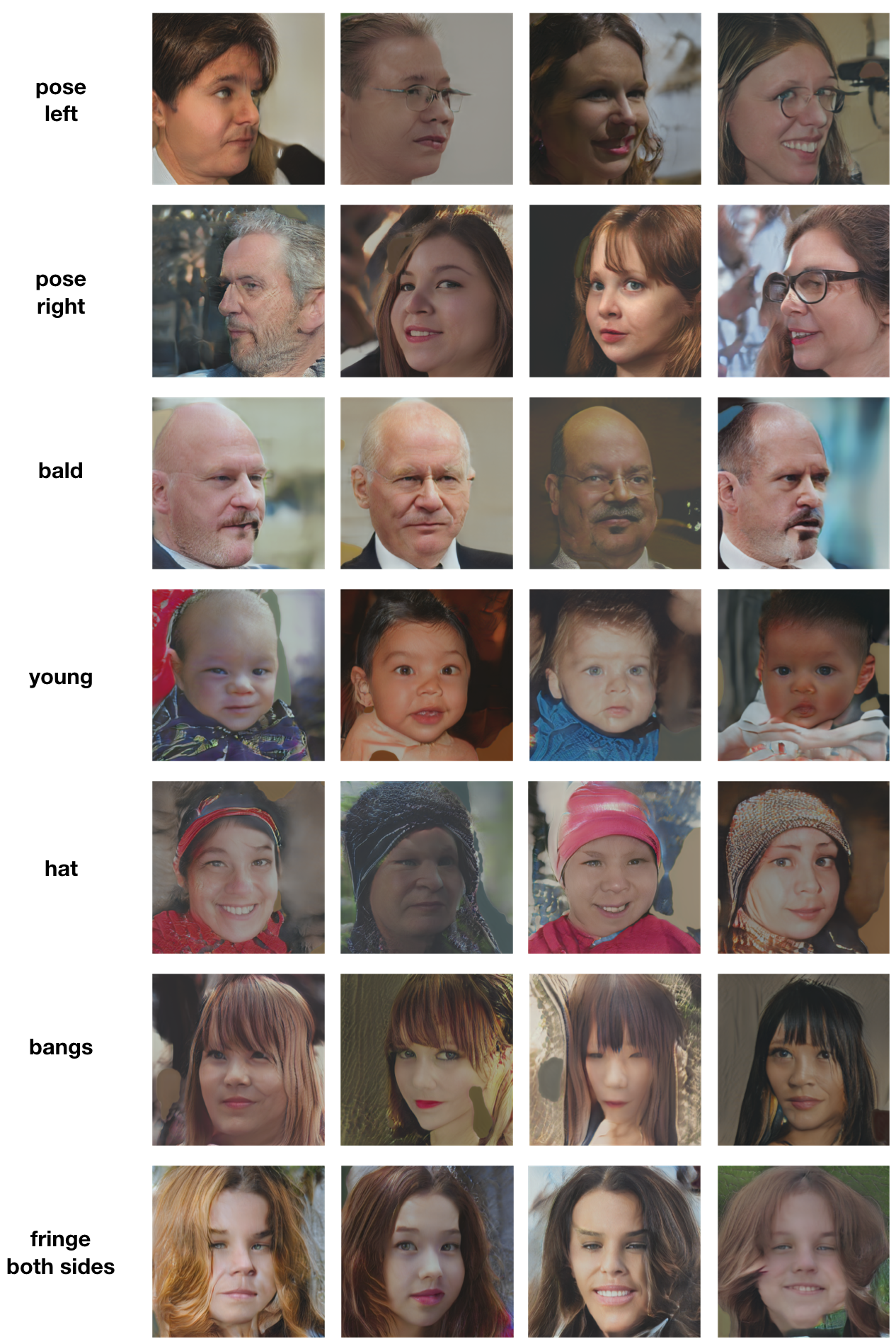}
  \caption{Each row depicts images synthesized from a different cluster in the representations of StyleGAN. The semantic attribute of each cluster is denoted on the left of the image. Note that besides primitive features like pose, the method captures several high-level attributes such as hat or bald.}
  \label{fig:stylegan_att}
}
\end{figure}

In this section, we validate the proposed method on a variety of architectures across different datasets. In particular, we utilize PGAN~\cite{karras2017progressive} on LSUN~\cite{yu2015lsun} and CelebA-HQ~\cite{karras2017progressive}, StyleGAN~\cite{karras2018style} on FFHQ and BigGAN~\cite{brock2019large} on Imagenet~\cite{russakovsky2015imagenet}. Our evaluation demonstrates that the proposed clustering works when trained in different objects, such as faces or cars. We verify that these clusters contain semantically-relevant images by showcasing state-of-the-art attribute classification results compared to four baselines.

\subsection{Experiments on faces}
To highlight the effect of our method on representations of different layers, we utilize PGAN trained on CelebA-HQ. Conditional image synthesis for binary attributes is presented in Figure \ref{fig:pgan_layers}. For this experiment, the representation space was separated into 2 clusters for each layer. The results highlight that different semantic features are captured in different layers. Indicatively, the first layer captures gender, while geometric and color features are encoded in the later layers.

The results of our method in Figure \ref{fig:stylegan_att} showcase that the representation forms multiple clusters based on attributes like pose, hair style and age.

\subsection{Experiments on objects}
\begin{figure*}
\centering{
  \includegraphics[width=0.7\textwidth]{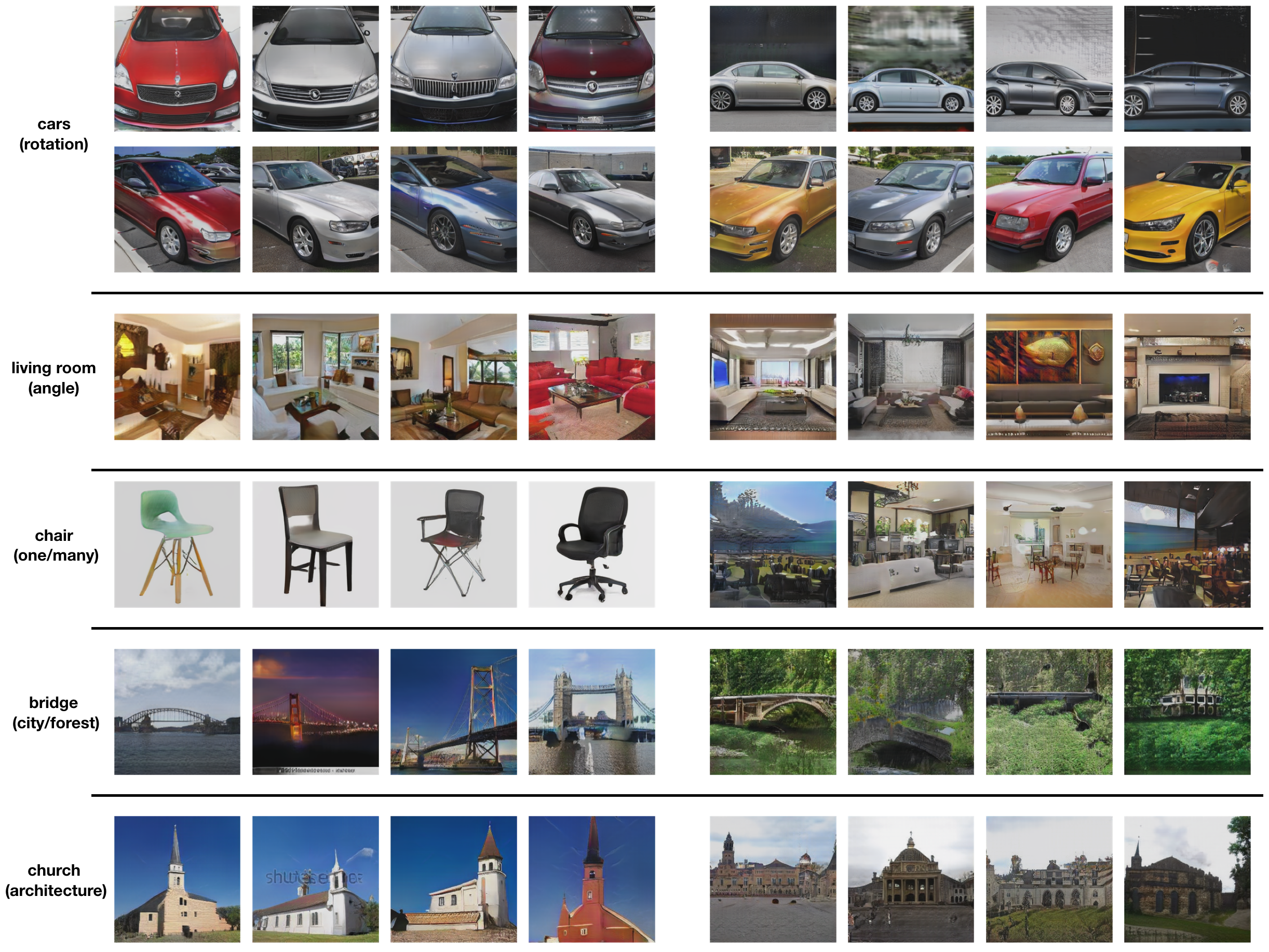}
  \caption{Synthesized images on LSUN with Progressive GAN (PGAN). The proposed approach can identify different sources of variation in each cluster, for instance rotation in the car, or multitude of objects in the chairs, background context in the bridges, and even architectural style in the churches. }
  \label{fig:pgan_lsun_objects}
  }
\end{figure*}

In addition to faces, we demonstrate in this section how our method generalizes on the object classes of LSUN. 
We notice that by using our method we obtain direct control of the rotation of cars, as well as other high level features of different classes. Most of the recovered clusters show considerable variation, e.g., one vs many chairs, the landscape behind a bridge and the architecture of the church. The results on Figure \ref{fig:pgan_lsun_objects} show that our method can introduce control of significant modes of variation without loss of photo-realism.

\subsection{Fine-grained attribute synthesis}

\begin{figure}
  \includegraphics[width=\linewidth]{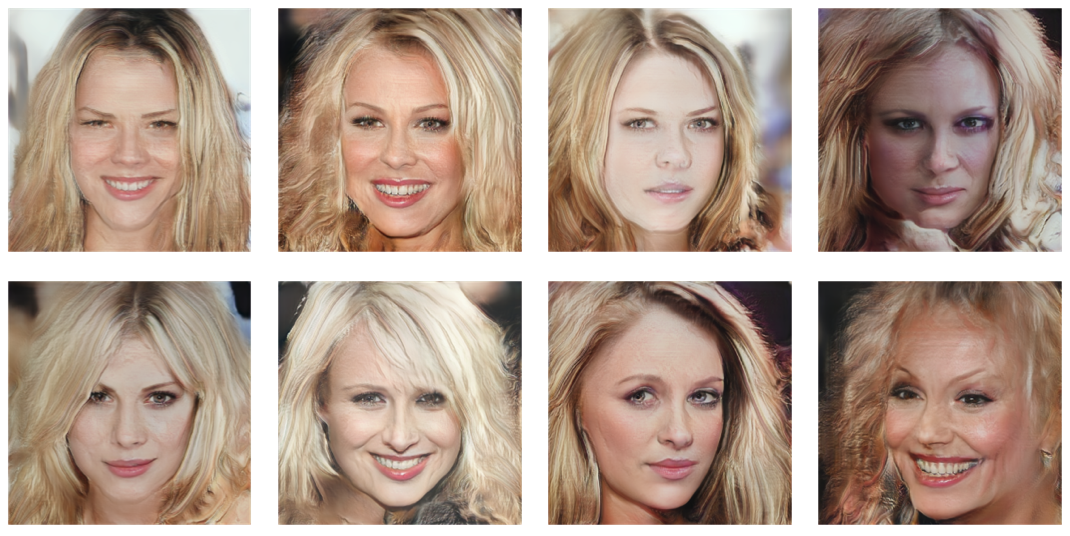}
  \caption{Fine-grained attribute synthesis for attributes female and blond hair.}
  \label{fig:pgan_hierarchical}
\end{figure}

As showcased in Figure \ref{fig:pgan_layers}, different semantic features are captured in different layers of PGAN. A logical extension would be to attempt to combine such features to form fine-grained attributes in a hierarchical manner. We indeed verify this assumption in Figure \ref{fig:pgan_hierarchical} where we sample blond female faces using PGAN on CelebA-HQ.
Forming the cluster for fine-grained attributes requires two steps. First,  we sample latent codes from the latent distribution, perform a forward pass and learn the clusters on the representation space of the first attribute (e.g., the first layer for gender). Then we perform a forward pass using only the samples of the specified cluster (e.g., female faces) onto the later layers (e.g., the fifth layer for hair tone). After clustering again, the resulting cluster will only correspond to blond female faces. The rest of the sampling procedure is trained using IMLE as discussed above.

\subsection{Generalization across classes}
\begin{figure*}[h!]
\centering{
  \includegraphics[width=0.7\linewidth]{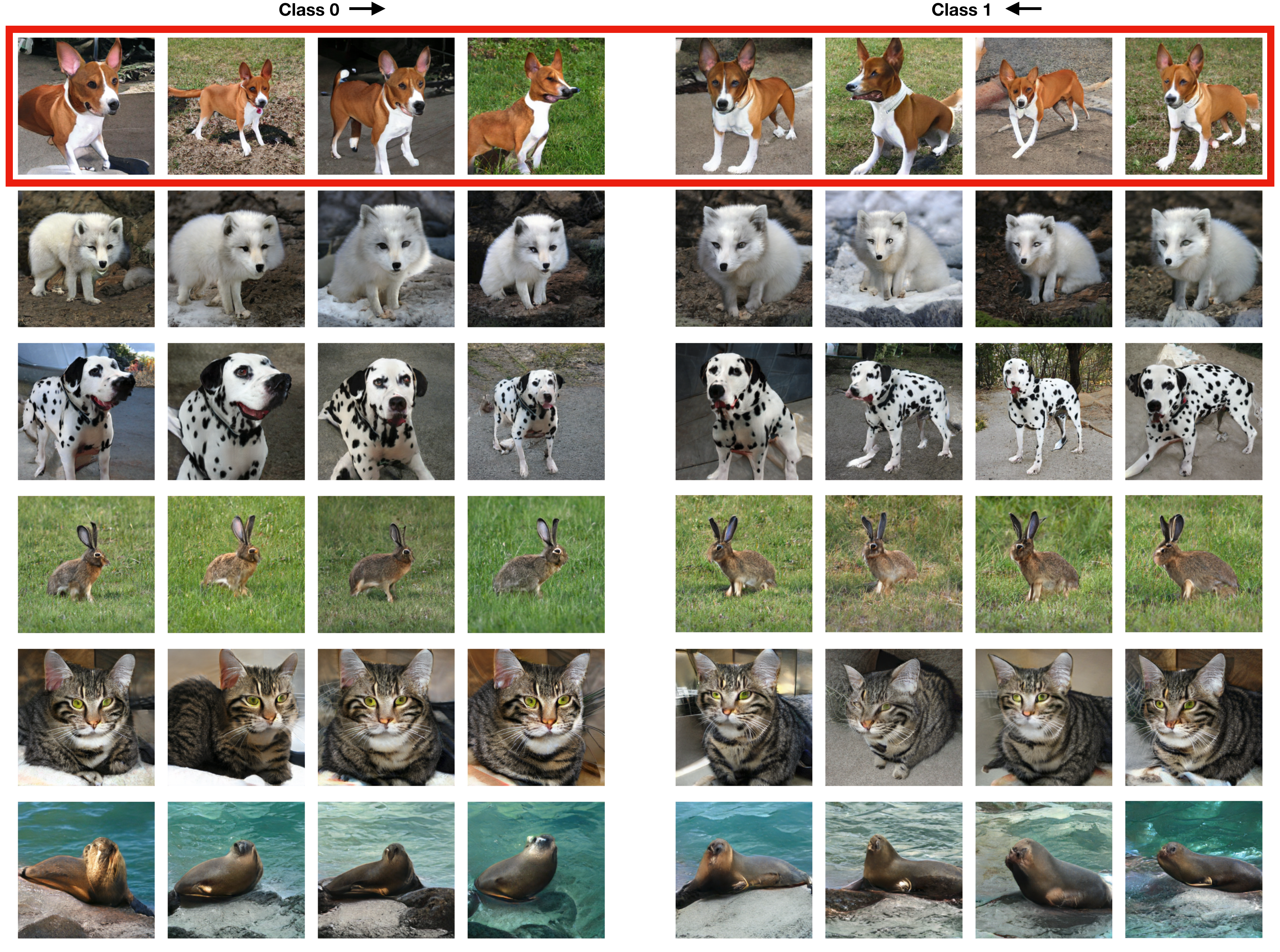}
  \caption{Generalization to different classes. The two mappings are learned in the representation space of the class depicted in the first row (i.e., red rectangle). Then, we demonstrate how the learned mappings transfer to other ImageNet classes without training them. Note that the transformation generalizes beyond other dog breeds, e.g., it applies to rabbits and seals.}
  \label{fig:across_classes}
  }
\end{figure*}

We further explore the generalization of the mappings that are learned from IMLE across different classes of objects. In particular, we are interested in determining whether a mapping $T_c$ that is responsible for a specific attribute for one class (e.g., pose of a dog) can be used on a different class (e.g., a cat) to facilitate the same attribute.
To this end, we utilize BigGAN~\cite{brock2019large} trained in Imagenet. BigGAN is already a conditional model, however it only allows for object class labels (e.g., dog breed). 
However, when our method is used in conjunction with the model, we can control generalizable geometric attributes like pose. In Figure \ref{fig:across_classes} we train the two mapping networks (one for each pose) on one object class (in the first row) and use them to sample images of different animals. The results highlight that the geometric features are encoded in the same layers for similar classes and hence, the same mapping can be used across different classes. This finding indicates that the generator learns to disentangle shape from appearance for classes with similar geometry (e.g., different cat breeds). In particular, the network learns generalisable low-level primitives for similar looking classes, e.g., pose. However, the same does not hold for higher-level class-specific attributes, such as type of car.

\subsection{Comparison to end-to-end training}
\begin{figure*}[h!]
  \includegraphics[width=\textwidth]{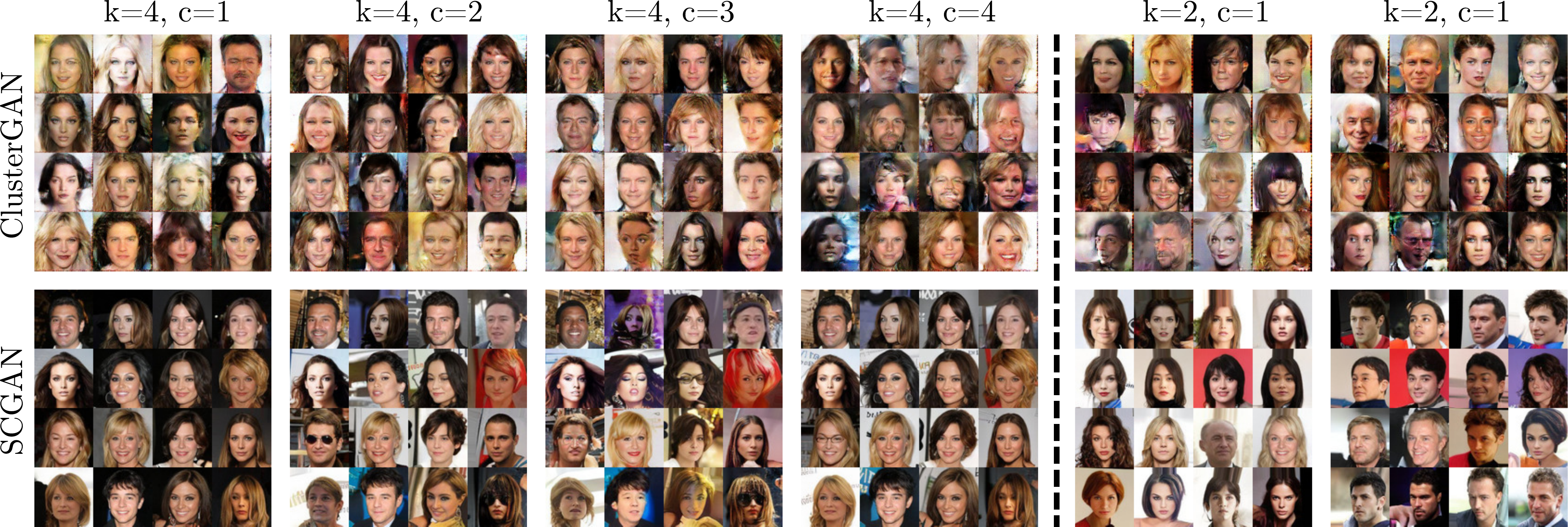}
  \caption{Results of SCGAN \cite{liu2020diverse} and ClusterGAN \cite{mukherjee2019clustergan} trained on CelebA for $k=2$ and $k=4$ total number of clusters. For SCGAN $k=2$, we notice that the separated attribute is gender entangled with pose, while for SCGAN $k=4$ there is no clear disentangled attributed except for the $c=1$ cluster, which captures face with dark backgrounds. On the other hand, the clusters produced by ClusterGAN do not demonstrate consistent attributes.}
  \label{fig:clustergan}
\end{figure*}

One of the advantages of the proposed method is that it can be used on any pretrained GAN, without the need for retraining the generator. 
However, we compare against SCGAN \cite{liu2020diverse} and ClusterGAN \cite{mukherjee2019clustergan} that learn a clustering that separates the latent space during training. We showcase results on CelebA for 2 and 4 clusters respectively in Figure \ref{fig:clustergan}. In the case of 2 clusters, the method mostly learns to separate female from male faces, entangled with pose. However, in the case of 4 clusters we do not notice a clear separation of semantic attributes other than image statistics (e.g., dark background) for SCGAN. We further quantify the inconsistency of attributes in each cluster in Table \ref{tab:attribute-predictions}. In particular, we classify hair color and gender in each cluster and present the percentage corresponding to `dark hair' and `female'. The results highlight that in some cases the distribution may even be almost uniform, indicating attribute inconsistency. Similarly, we calculate the yaw of the faces. The inconsistency in pose is demonstrated by the large standard deviation.

\begin{table}[b!]
\centering
\resizebox{\linewidth}{!}{%
\begin{tabular}{l|llll}
Model & Hair & Gender & Yaw (deg)  \\ \hline
SCGAN, k=4, c=1 & 79\% & 94\% & -6.79 $\pm$ 5.19 \\
SCGAN, k=4, c=2 & 85\% & 45\% & -13.4 $\pm$ 4.93 \\
SCGAN, k=4, c=3 & 65\% & 86\% & 11.46 $\pm$ 7.51  \\
SCGAN, k=4, c=4 & 73\% & 95\% & -9.04 $\pm$ 5.72 \\
\hline
SCGAN, k=2, c=1 & 72\%  & 92\% & -2.64 $\pm$ 5.86  \\
SCGAN, k=2, c=2 & 82\% & 35\% & -24.30 $\pm$ 7.59 \\
\hline
ClusterGAN, k=4, c=1 & 69\% & 89\% & -1.72 $\pm$ 5.88  \\
ClusterGAN, k=4, c=2  & 54\% & 76\%  & -2.42 $\pm$ 6.49 \\
ClusterGAN, k=4, c=3 & 67\% & 73\% & -14.58 $\pm$ 7.81 \\
ClusterGAN, k=4, c=4 & 67\% & 75\% & 7.50 $\pm$ 7.60  \\
\hline
ClusterGAN, k=2, c=1 & 56\% & 80\% & -11.59 $\pm$ 7.77 \\
ClusterGAN, k=2, c=2 & 72\% & 69\% & 5.30 $\pm$ 9.5 
\end{tabular}%
}
\caption{Attribute predictions for the images of each cluster for ClusterGAN~\cite{mukherjee2019clustergan} and SCGAN~\cite{liu2020diverse}, trained for $2$ and $4$ clusters. For the attribute `hair' we report the percentage of `dark' hair (classified as `brown' or `black'). For the attribute of gender we report the percentage of faces classified as `female'. For `yaw', we report both the mean and standard deviation of the degrees.}
\label{tab:attribute-predictions}
\end{table}

\subsection{Quantitative comparison} 
To evaluate our method quantitatively, we generate samples for the attributes `yaw' and `gender' and evaluate them using Microsoft Azure\footnote{\url{https://azure.microsoft.com/en-gb/services/cognitive-services/face/}}. 
We compare against GANSpace \cite{harkonen2020ganspace} and SeFA \cite{shen2020closed} by identifying the above attributes in their basis of interpretable directions. 
Similarly, we also compare against ClusterGAN and SCGAN by identifying the clusters where each studied attribute is more prevalent.
The results in Table \ref{tab:classification-scores} show that the images generated using the proposed method consistently contain the target attributes. On the other hand, ClusterGAN and SCGAN are not able to find clusters that separate the pose. 
\begin{table}[]
\centering
\begin{tabular}{l|llllll}
& Gender & Yaw\\ \hline
ClusterGAN  & 51\%  & 24\%\\
SCGAN  & 95\%  & 58\% \\
GANSpace  & 98\%  & 89\% \\
SeFA  & 98\% & 94\% \\ \hline
Ours  & \textbf{100\%}  & \textbf{95\%}
\end{tabular}%
\caption{Classification accuracy for multiple attributes using the baseline methods and ours. For gender, we sample images with the `female' attribute and for yaw with the `pose right' (at least 10 degrees from frontal). If a face is not found in the generated image, we deem it to be misclassified.}
\label{tab:classification-scores}
\end{table}

\subsection{Implementation details}
The mapping networks consist of $3$ fully connected layers without biases, as well as batch-normalization between each layer \footnote{\url{https://www.math.ias.edu/~ke.li/projects/imle/}}. The networks were optimized using Adam~\cite{kingma2014adam} on Pytorch~\cite{pytorch}. We train each model for $400$ epochs on a Titan X GPU with $12$ GB in less than an hour. Both k-means and the nearest neighbour algorithms are implemented using FAISS \cite{JDH17}.
To enable reproducibility of our work, we utilize open-source code for the pretrained GANs (the links can be found in the supplementary material).

%% file: sections/limitations.tex
\section{Limitations and broader impact}
\paragraph{Limitations:}The qualitative results in the previous section highlight the efficacy of the proposed method in conditioning unconditional GANs. However, the generated attributes can be entangled and are dependent on the variation present in the training set (e.g., bald people are always male in Figure \ref{fig:stylegan_att}). It should be noted that similar limitations are faced by most unsupervised methods (e.g., \cite{harkonen2020ganspace}). Furthermore, the number of clusters used for k-means has an effect on the resulting attributes. In this work, we treat the number of clusters as a hyper-parameter but there are several heuristics in the literature that deal with this issue (e.g., elbow method or eigengap for subspace clustering\cite{von2007tutorial}). However, calculating the number of clusters is beyond the exploratory purpose of this work. Lastly, since the mapping learned by IMLE is approximate, the generation quality of the samples is not always similar to the ones sampled from the latent distribution, which is a trade-off for controllable synthesis.
\paragraph{Broader impact:}
Our method is built on top of a pretrained GAN. As such, it inherits all the biases of the training data used to train the GAN, e.g., issues with CelebA-HQ. Our method can be used to control the low-level features (e.g., pose) for on-demand generation. If the dataset includes biases, those could be reflected in the clusters and invariably in the on-demand generation. On the other hand, our method can be viewed as a tool for investigating such biases, since the clusters will reflect the primary variations of the dataset. We also emphasize that the high-fidelity GANs we utilize~\cite{brock2019large, karras2018style, karras2017progressive} are publicly available. As the generation quality improves further, we believe methods like ours can be used as a `semantic debugging tool' to uncover the biases of the GAN model. Thus, we believe our work aids towards transparency and explainability in generative models.

%% file: sections/conclusion.tex
\section{Conclusion}
\label{sec:cluster_gan_conclusion}

In this work, we introduce a method for controllable generation using unconditional GANs. The proposed method focuses on learning semantic attributes without supervision and conditioning the GAN generator using such attributes.  Our method is lightweight and can work on top of any GAN generator as demontrated by our experiments with three strong-performing generators, i.e., Progressive GAN, StyleGAN and BigGAN. We explore how those semantic attributes differ across classes and even illustrate how learned attributes in one class can transfer to different classes. A future step would be to explore automatic selection of the number of clusters.    